\documentclass[sigconf]{acmart}

\usepackage{multirow}
\usepackage{subcaption}
\usepackage{enumitem}
\usepackage{balance}
\usepackage{multicol}
\usepackage{stfloats}
\usepackage{booktabs} 
\usepackage{multirow} 
\usepackage{multicol} 

\usepackage{algorithm}
\usepackage{algpseudocode}
 
\AtBeginDocument{%
  }

\copyrightyear{2024}
\acmYear{2024}
\setcopyright{acmlicensed}\acmConference[ICMR '24]{Proceedings of the 2024 International Conference on Multimedia Retrieval}{June 10--14, 2024}{Phuket, Thailand}
\acmBooktitle{Proceedings of the 2024 International Conference on Multimedia Retrieval (ICMR '24), June 10--14, 2024, Phuket, Thailand}
\acmDOI{10.1145/3652583.3658073}
\acmISBN{979-8-4007-0619-6/24/06}

\begin{document}

\title{NeurNCD: Novel Class Discovery via Implicit Neural Representation}


\author{Junming Wang}
\email{jmwang@cs.hku.hk}
\authornote{Corresponding author.}
\affiliation{%
  \institution{The University of Hong Kong}
  \streetaddress{}
  \city{Hong Kong SAR}
  \country{China}}

\author{Yi Shi}
\email{21120237@bjtu.edu.cn}
\affiliation{%
  \institution{Beijing Jiaotong University}
  \streetaddress{}
  \city{Beijing}
  \country{China}}

\renewcommand{\shortauthors}{Junming Wang \& Yi Shi}

%

\begin{teaserfigure}
  \includegraphics[width=\textwidth]{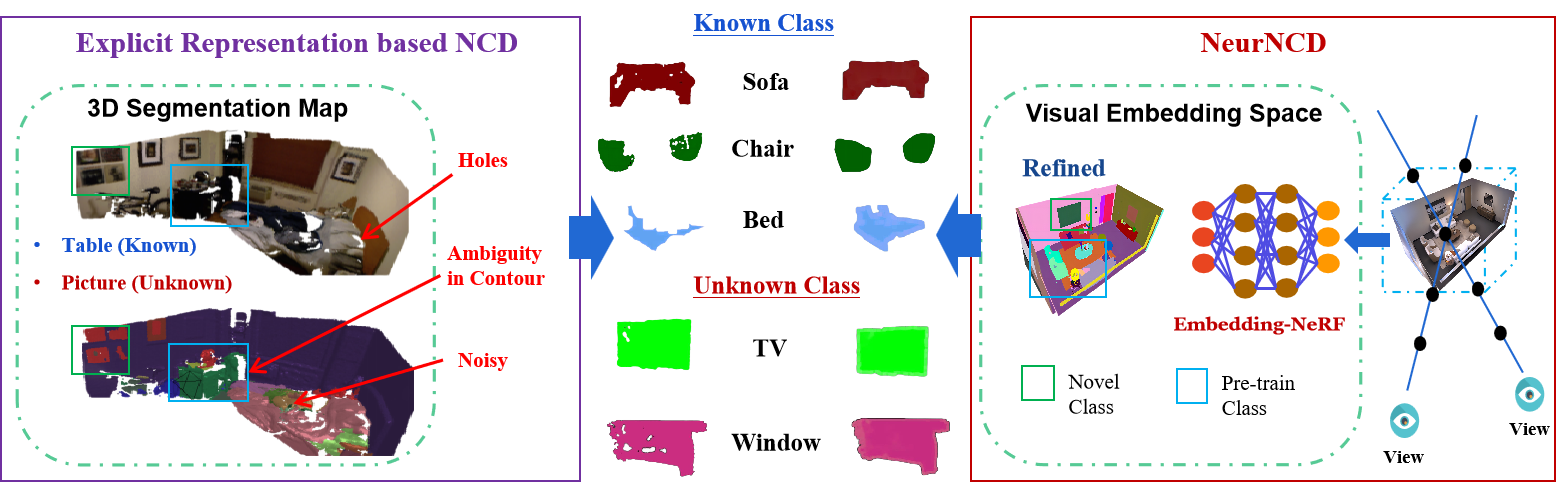}
  \caption{NeurNCD leverages implicit neural representation, replacing traditional explicit 3D segmentation maps\cite{nakajima2019incremental}, to enhance the accuracy of novel class discovery. Specifically, the meticulously designed Embedding-NeRF model employs KL divergence, achieving the transfer and association of 2D–3D features while producing semantic embedding and entropy by aggregating information from multiple views. Then by integrating with other key components, i.e., feature query, feature modulation and clustering, to ultimately reconstruct refined, low-noise, and hole-free images and 3D structures.}
  \label{fig:head}
\end{teaserfigure}

\begin{abstract}
Discovering novel classes in open-world settings is crucial for real-world applications. Traditional explicit representations, such as object descriptors or 3D segmentation maps, are constrained by their discrete, hole-prone, and noisy nature, which hinders accurate novel class discovery. To address these challenges, we introduce NeurNCD, the first versatile and data-efficient framework for novel class discovery that employs the meticulously designed Embedding-NeRF model combined with KL divergence as a substitute for traditional explicit 3D segmentation maps to aggregate semantic embedding and entropy in visual embedding space. NeurNCD also integrates several key components, including feature query, feature modulation and clustering, facilitating efficient feature augmentation and information exchange between the pre-trained semantic segmentation network and implicit neural representations. As a result, our framework achieves superior segmentation performance in both open and closed-world settings without relying on densely labelled datasets for supervised training or human interaction to generate sparse label supervision. Extensive experiments demonstrate that our method significantly outperforms state-of-the-art approaches on the NYUv2 and Replica datasets.
\end{abstract}




\maketitle

\section{Introduction}
The swift advancements in computer vision and robotics have transitioned from "Supervised AI" to "Embodied AI," whereby AI algorithms and agents can learn through interactions with their environment, adopting a human-like egocentric perspective. However, most existing perception algorithms\cite{gupta2015indoor,chen2018encoder,seichter2021efficient,wang2022card,wang2024agrnav} operate in a closed-world setting and are trained to segment a limited number of semantic classes, which do not adequately address the needs of Embodied AI applications in constantly changing and open environments. In open-world perception scenarios, novel classes continually emerge, but a perception model trained on a limited number of semantic classes may either treat a novel class as background or misclassify it as one of the known objects\cite{yang2020convolutional}. Ideally, an Embodied AI system should discover and incrementally learn to recognize novel classes through interactions with the environment. This challenge, known as novel class discovery, has garnered significant interest within the research community and has crucial real-world implications.

Humans are adept at recognizing novel classes by identifying consistent features within their surroundings. Drawing on this observation,  \citeauthor{nakajima2019incremental}  proposed a method \cite{nakajima2019incremental} for aggregating spatially consistent features by explicitly constructing a 3D segmentation map and clustering to discover novel classes (in Fig.\ref{fig:head}). Although this method demonstrates the feasibility of the novel class discovery task and presents a solution framework, it suffers from several limitations related to noise, hole-prone, overlapping, and mapping errors, which significantly impact the accuracy of discovery of novel classes. Furthermore, the method introduces manually designed update strategies for continuously updating semantic features and entropy, leading to additional hyperparameters and suboptimal strategies.

Recently, implicit neural representations have gained considerable attention due to their exceptional performance in 3D scene modelling from novel viewpoints. Thus, we investigate whether implicit neural representations can replace traditional explicit 3D segmentation maps to enhance novel class discovery accuracy. In contrast to a conventional point cloud or voxel-based 3D map, implicit neural representations utilize a multilayer perceptron (MLP) to represent a 3D map. They can reconstruct low-noise and hole-free 3D structures by aggregating information from different perspectives. Moreover, implicit neural representations have been employed to aggregate semantic labels\cite{zhi2021place}, achieving state-of-the-art performance. However, this method relies on dense annotation and cannot discover novel classes.

The above limitations show that implementing implicit neural representations for discovering novel classes demands addressing two significant challenges. The first challenge arises from the fundamental differences between implicit and explicit representations. Explicit representations (e.g., point clouds or voxels) enable direct operations like clustering to discover novel classes, whereas implicit representations require information extraction through queries for the same purpose. Consequently, it is crucial to devise a novel class discovery framework that accommodates the unique operational mode of implicit representations while considering complex factors, i.e., query-based information extraction.

The second challenge involves developing suitable methods for semantic feature aggregation and an update strategy tailored specifically for novel class discovery. Although NeRF's inherent feature aggregation capabilities allow for direct implementation of semantic feature aggregation and updates through error backpropagation, such as Semantic NeRF\cite{zhi2021place} employs softmax loss for aggregating semantic labels. However, this approach is unsuitable for novel class discovery, as softmax loss generally categorizes unknown classes as background classes\cite{yang2020convolutional}. This challenge highlights the necessity for designing an implicit representation specifically adapted for novel class discovery, potentially involving the exploration of new loss functions and the enhancement of feature extraction methodologies to overcome these limitations.

Tackling the challenges mentioned earlier, we present NeurNCD, the first versatile and data-efficient approach for novel class discovery using neural radiance fields and feature modulation, applicable to both "open set" and "closed set" scenarios. 

In general, the main contributions of this work are:
\begin{itemize}[noitemsep,topsep=0pt]
\item We address the challenge of discovering novel classes in implicit neural representation tasks by proposing a novel framework named NeurNCD. This method is versatile and data-efficient, thanks to the advantages of implicit representation in terms of refinement, low noise and continuity, the accuracy of new class discovery is significantly improved.
\item Our method introduces the key component, Embedding-NeRF, which is specifically designed for novel class discovery tasks, and enables the replacement of traditional explicitly constructed 3D segmentation maps. Through the minimization of the Kullback-Leibler (KL) divergence, it generates semantic embeddings and entropy, thus bestowing a considerable advantage in the realm of novel class discovery.
\item Extensive experiments demonstrate that our method outperforms state-of-the-art approaches on NYUv2 and Replica datasets. The design of each component is supported by comprehensive experimental validation and extensive ablation investigations.
\end{itemize}

\section{Related work}
\subsection{Semantic Segmentation}
As a fundamental task in computer vision, semantic segmentation, which seeks to predict semantic labels for every pixel in an image, has received much attention. In recent years, substantial progress has been achieved in the field of supervised semantic segmentation\cite{li2021ctnet,liu2018deep}, however, such work is ``labour-intensive" and appears to be at a loss when confronted with new environments or unknown classes.

In order to remove the dependence on annotations, Unsupervised semantic segmentation has caught the interest of researchers because it can reduce the amount of pixel-level annotations needed for semantic segmentation while also discovering novel classes.  \citeauthor{nakajima2019incremental} were one of the first works to discover novel classes, they rely on superpixel segmentation, mapping, and clustering to identify object categories. \citeauthor{9874976} shows a ready-to-deploy continuous learning approach for semantic segmentation that does not require any prior knowledge of the scene or any external supervision and can simultaneously retain the knowledge of previously seen environments while integrating new knowledge. In order to deploy the semantic segmentation model on the robot, \citeauthor{seichter2021efficient} proposed ESANet, which is an efficient and robust RGB-D segmentation approach that can be optimized to a high degree using NVIDIA TensorRT \cite{vanholder2016efficient}. They evaluated ESANet on the common indoor datasets NYUv2 and SUNRGB-D, and the results demonstrated that the method achieves state-of-the-art performance while enabling faster inference.

\begin{figure*}[t]
  \centering
  \includegraphics[width=\linewidth]{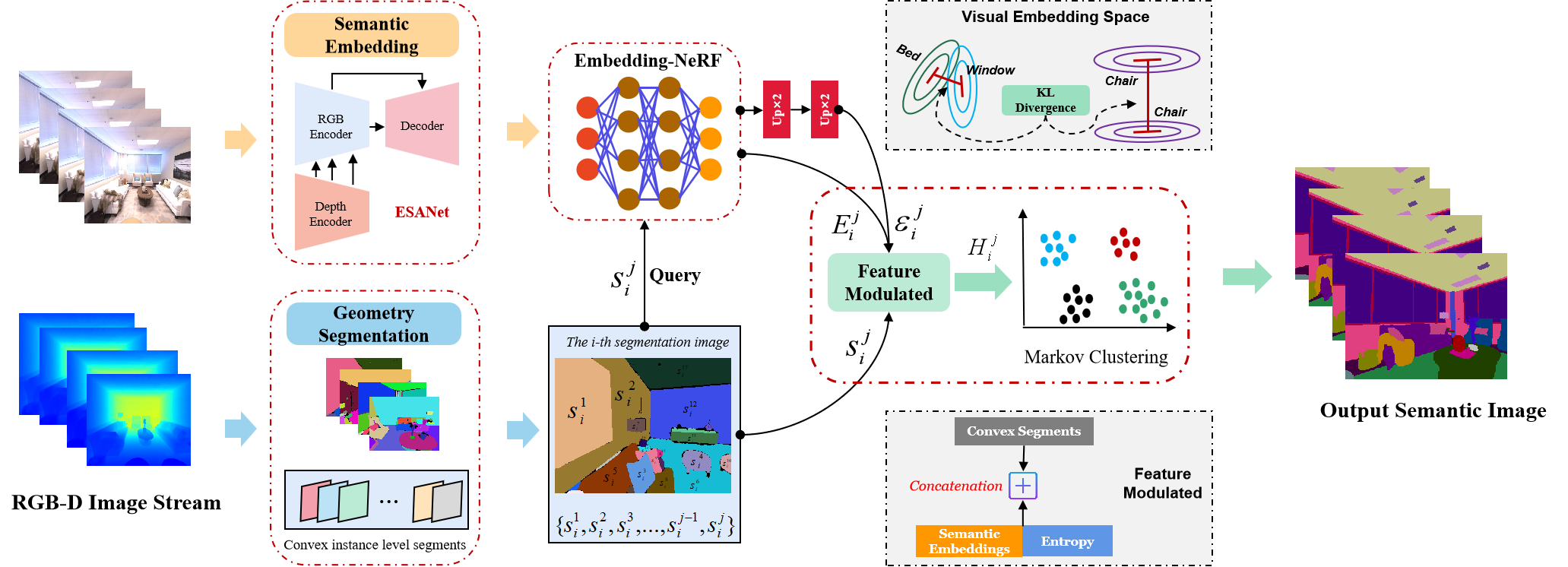}
  \caption{An overview of our method. For input RGB-D image, the pre-trained semantic segmentation network $f_{\theta}$ to extract semantic embedding, and then use our proposed Embedding-NeRF $F_{\theta}(P)$ model to generate globally consistent semantic embedding and entropy. Meanwhile, we leverge the geometric segmentation $\mathcal{F} $ to obtain a set of convex sub-instance-level segments $\left \{ s_i^1,s_i^2,s_i^3,...,s_i^{j-1},s_i^j \right \}$. These segments will then go to the output of Embedding-NeRF to query the corresponding semantic embedding $E_{i}^j$ and entropy $\varepsilon_i^j$. Finally, in the feature modulated, the three parts were concatenated and clustered which can obtain the final semantic segmentation results (including known classes and novel classes).}
   \label{fig:onecol}
\end{figure*}

\subsection{Radiance Field-based Scene Representations}

Our work on discovering novel classes and unsupervised semantic segmentation build on neural radiance fields (NeRF) \cite{mildenhall2021nerf}, which represent a scene using a multi-layer perceptron (MLP) that maps positions and directions to densities and radiances. The following work\cite{xu2022point,muller2022instant,wang2022r2l,chen2022structnerf,fu2022panoptic,wu2022dof,Barron_2022_CVPR,Turki_2022_CVPR,guo2022neural} improve NeRF for faster training and inference and more realistic rendering. Using MLP or explicit feature grids, these radiance field-based scene representations achieve unprecedented novel view synthesis effects. Considering Semantic Segmentation in the Neural Radiation Field, NeSF\cite{vora2021nesf}, a method for simultaneous 3D scene reconstruction and semantic segmentation from posed 2D images, is demonstrated by Suhani Vora et al. Their approach, which is based on NeRF, is trained entirely on posed 2D RGB images and semantic mappings. Their method creates a dense semantic segmentation field during inference that can be queried directly in 3D or used to produce 2D semantic maps from novel camera postures, but their method is a supervised method the same as \cite{zhi2021place} and \cite{zhi2022ilabel}. At the same time, their method only verified that NeRF has a strong ability in the low-dimensional image or semantic rendering, but did not research the performance of NeRF fusion and rendering of higher-dimensional image embeddings.

\subsection{Novel Class Discovery and Clustering}
For novel class discovery, \citeauthor{zhao2022novel} proposed a method to discover novel classes with the help of a saliency detection model and use an entropy-based uncertainty modelling and self-training (EUMS) framework to overcome noisy pseudo-labels, further improving the model's performance on the novel classes. But their method can only segment and discover a limited number of salient categories, while our method can segment all categories in the entire indoor scene with the help of implicit representations.For classification, this can be understood as a two-part problem. First, high-dimensional descriptors for the items in question have to be found. Then, a clustering algorithm groups similar descriptors together. The established approach in representation learning is to learn a single good descriptor that can be clustered with KNN or k-means \cite{hamerly2003learning}. K-means can be used with mini-batches and is differentiable, fast, and easy to implement. However, we argue that there are two big disadvantages: it requires a priori knowledge of the number of clusters k and only works in the space of a single descriptor. An alternative graph-based clustering algorithm like Markov clustering \cite{ye2022deep} performs effective random walks for unsupervised clustering without pre-defined cluster numbers.

\section{Method}
In this section, we present NeurNCD, a method specifically designed for novel class discovery comprised of several key components. An overview of our approach is shown in Figure 2. 

Firstly, to enhance the accuracy of novel class discovery and replace traditional explicit 3D segmentation maps, we delve into the visual embedding space to decode valuable features for semantic classes, namely, semantic embedding and entropy. To achieve this, we introduce the \textbf{\textit{Embedding-NeRF model}}, which employs KL divergence\cite{yang2020convolutional} to facilitate the migration and aggregation of 2D-3D features and then generates the above features. Meanwhile, minimizing KL divergence loss achieves multiple goals, such as reducing distances between genuine prototypes, increasing distances among incorrect prototypes, and effectively capturing unknown class feature distributions — an advantage absent in cross-entropy loss. Moreover, Embedding-NeRF also generates entropy as an uncertainty measure for semantic classes, offering robust supplementary information for novel class discovery (Section 3.1). Upon acquiring globally consistent semantic embeddings and entropy, we adopt the \textbf{\textit{Geometric Segmentation}} method from \cite{nakajima2019incremental} to segment depth images, dividing both known and new classes into a set of convex sub-instance-level segments (Section 3.2). However, solely relying on segments' geometric features is not enough for novel class discovery due to the absence of globally consistent semantic information. As a result, we fuse semantic embeddings and entropy with these segments to enhance the features. Specifically, we query the corresponding semantic embeddings and entropy for the segments from the Embedding-NeRF output and concatenate them during the \textbf{\textit{Feature Modulation}}, ensuring that segments of the same class share identical semantic embeddings and entropy (Section 3.3). Ultimately, we achieve known class segmentation and novel class discovery using \textbf{\textit{Markov clustering}} based on cosine similarity (Section 3.4).

\begin{figure*}
  \centering  
   \includegraphics[width=\linewidth]{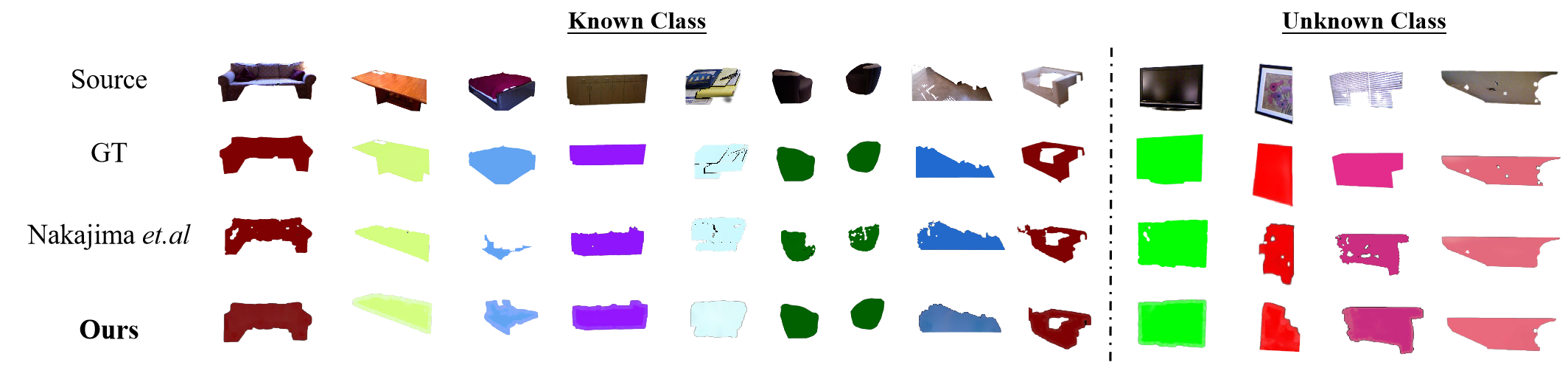}
   \caption{Quantitative results for known and unknown classes in the NYUv2 dataset. With the powerful feature propagation and fusion capabilities of Embedding-NeRF, our method is very complete and smooth for each class segmentation, the baseline method relies on geometric segmentation results, and there are segmentation errors or incomplete phenomena.}
   \label{fig:onecol}
\end{figure*}

\subsection{Embedding-NeRF}
\begin{figure}[!htb]
  \centering
    \includegraphics[width=\linewidth ]{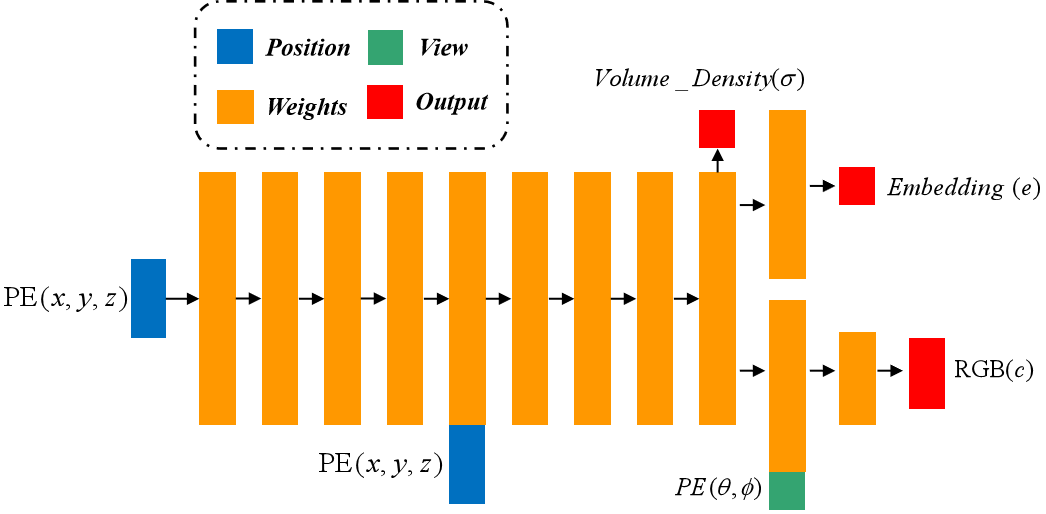}
   \caption{\textbf{Embedding-NeRF} 3D position $(x,y,z)$  and viewing direction $(\theta,\phi )$ are fed into the network after positional encoding (PE). Volume density $ \sigma $ and semantic embedding $e$ are functions of 3D position while colours $c$ additionally depend on viewing direction. }
   \label{fig:onecol}
\end{figure}

In the context of novel class discovery tasks, semantic embedding and entropy \cite{wang2022multimodal,liu2022cmx,chen2020bi} play a crucial role in enhancing the accuracy and robustness of semantic classes. Consequently, we draw inspiration from the burgeoning field of implicit neural representation and propose the Embedding-NeRF model as a substitute for the traditional explicit 3D segmentation map, and generate these two features using this model.

\noindent\textbf{Extract Semantic Embedding as Input}. Utilizing a pre-trained semantic segmentation network $f_{\theta}$ with parameters ${\theta}$, we extract the semantic embedding of each set of RGB-D images, denoted as $\Theta = (I_{i}^{RGB}, I_{i}^{D})$, where $I_{i}^{RGB}$ and $I_{i}^{D}$ correspond to an RGB image and its associated Depth image, respectively. Specifically, applying the segmentation network to the RGB-D datasets, we remove the network's classification layer and use the feature layer preceding the softmax layer as the input for Embedding-NeRF, denoted as $E_i$, a high-dimensional vector with dimensions of $N \times H \times W \times S$, where $S=37$. Consequently, the semantic embedding extraction process can be expressed as:
\begin{equation}
E_{i} = f_{\theta } (\Theta )=f_{\theta }((I_{i}^{RGB},I_{i}^{D}))
\end{equation}

\noindent\textbf{Embedding-NeRF}. NeRF \cite{mildenhall2021nerf} approximates volume rendering by numerical quadrature with hierarchical stratified sampling to determine the color of a single pixel. Within one hierarchy, if $r(t) = o$ + $td$ is the ray emitted from the centre of projection of camera space through a given pixel, traversing between near and far bounds($t_n$ and $t_f$),  then for selected $K$ random quadrature points ${\left \{ t_k \right \} }_{K}^{k=1}  $ between $t_n$ and $t_f$, the approximated expected colour is given by:
\begin{equation}
\hat{C}(r) = \sum_{k=1}^{K}\hat{T}(t_{k})\alpha (\sigma (t_{k})\delta _{k})c(t_k) 
\end{equation}
where \begin{equation}
\hat{T}(t_{k}) = exp(-\sum_{k^{'}=1}^{k-1}\sigma (t_k)\delta _k ) 
\end{equation}
where $\alpha (x) = 1-exp(-x)$ and $\delta _{k} = t_{k+1}-t_{k}$ is the distance between adjacent sample points.

We now show how to extend NeRF to jointly encode appearance, geometry and embedding. As shown in Figure 4, we augment the original NeRF by appending an embedding renderer before injecting viewing directions into the MLP.
\begin{equation}
F_{\theta}(P) = (c,e,\sigma)
\end{equation}
where $F_{\theta}$ is a MLP parameterised by $\theta$; $c,e$ and $\sigma$ are the radiance, embedding logits and volume density at the 3D position $P = (x,y,z)$, respectively. The approximated expected embedding logits $\hat{E}(r)$ of a given pixel in the image plane can be written as:
\begin{equation}
\hat{E}(r) = \sum_{k=1}^{K}\hat{T}(t_{k})\alpha (\sigma (t_{k})\delta _{k})e(t_k) 
\end{equation}where $\hat{T}(t_{k})$, $\alpha (x)$ and  $\delta _{k}$  are consistent with the definitions in NeRF. 

Embedding logits can then be transformed into multi-class probabilities through a softmax normalisation layer.
We train the whole network from scratch under photometric loss $L_p$ and embedding loss $L_e$:
\begin{equation}
L_p = \sum_{r\in R}\left [ \left \| \hat{C}_{c}(r)-{C}(r)  \right \|_{2}^{2} + \left \| \hat{C}_{f}(r)-{C}(r)  \right \|_{2}^{2}  \right ] 
\end{equation}

\begin{equation}
L_e = D_{KL}\left(E(r)||\hat{E}(r) \right) = \sum_{n=1}^{N}E(r_n)\log \frac{E(r_n)}{\hat{E}(r_n)} 
\end{equation}
where R are the sampled rays within a training batch, and $C(r), \hat{C}_{c}(r)$ and $\hat{C}_{f}(r)$ are the ground truth, coarse volume predicted and fine volume predicted RGB colours for ray $r$.
$L_e$ is chosen as a KL-divergence loss \cite{yang2020convolutional} to encourage the rendered embedding  $\hat{E}(r)$ to be consistent with the embeddings extracted by the pre-trained model $E(r)$, whether these are ground-truth, noisy or partial observations. Hence the total training loss L is:
\begin{equation}
L = L_p + \lambda L_e 
\end{equation}

\begin{table*}
\caption{Quantitative comparison on the NYUv2 dataset. Supervised methods, unsupervised methods versus our methods.}
\label{tab:apis}
\centering
\begin{tabular}{@{}l|ccccccccc|cccc|c@{}}
\toprule
\multirow{2}{*}{\textbf{Method}} & \multicolumn{9}{c|}{\textbf{Classes in Training Dataset}} & \multicolumn{4}{c|}{\textbf{Novel Classes}} & \multirow{2}{*}{\textbf{mIoU}} \\ \cmidrule(l){2-14} 
                                 & \textbf{Bed} & \textbf{Book} & \textbf{Chair} & \textbf{Floor} & \textbf{Furn.} & \textbf{Obj.} & \textbf{Sofa} & \textbf{Table} & \textbf{Wall} & \textbf{Ceil.} & \textbf{Pict.} & \textbf{TV} & \textbf{Window} & \\ \midrule
\cite{seichter2021efficient}     & 49.62        & 25.08         & 40.67          & 49.85          & 53.74          & 21.11         & 42.55         & 43.36         & 55.62         & -              & -              & -            & -              & - \\
\cite{nakajima2018fast}          & 62.82        & 27.27         & 42.56          & 68.43          & 44.62          & 24.63         & 45.04         & 42.30         & 26.82         & -              & -              & -            & -              & - \\
\cite{tateno2015real} + 3D Map   & 62.80        & 23.96         & 33.10          & 63.41          & 50.58          & 27.28         & 58.68         & 40.23         & 54.53         & 31.42          & 19.37          & 43.98        & 31.30          & 41.59         \\
\cite{nakajima2019incremental}   & 64.22        & 22.28         & 41.79          & 67.38          & 56.15          & 28.61         & 49.31         & 40.95         & 63.18         & 29.30          & \textbf{28.69} & 52.20        & 53.92          & 46.05         \\ \midrule
\textbf{NeurNCD}                             & \textbf{69.23} & \textbf{29.82} & \textbf{58.63} & \textbf{69.67} & \textbf{60.11} & \textbf{32.18} & \textbf{58.86} & \textbf{48.25} & \textbf{69.28} & \textbf{31.92} & 25.59          & \textbf{59.38} & \textbf{53.95} & \textbf{51.29} \\ \bottomrule
\end{tabular}
\end{table*}

In addition, since entropy can reflect the uncertainty of each semantic class, which is of great benefit for novel class discovery, so we use Embedding-NeRF to generate it from the visual embedding space. Specifically, after the semantic embedding $E_i$ obtained by fusion is sent to the two upsampling modules, the entropy $\mathcal{U}_{i}^o$ are obtained. In current frame $i$, the entropy $\varepsilon_i  \in \mathbb{R}$ is computed as follows:
\begin{equation}
\varepsilon_i = -\sum_{o\in \mathcal{O} }{\mathcal{U}_{i}^olog \mathcal{U}_i^o} 
\end{equation}
where $\mathcal{U}_{i}^o \in \mathbb{R}$ is the probability for the $o$ th class  in  $i$ th frame.

\subsection{Geometric Segmentation}

When the two types of features were generated by Embedding-NeRF, inspired by the work of \cite{furrer2018incremental},  We use geometric segmentation to segment the input depth image into a set of convex sub-instance-level segments, this process also can be regarded as an extract translation/rotation-invariant and noise-robust geometric features for known classes and novel classes.  We denote geometric segmentation as $\mathcal{F} $ and apply it to the depth image, where $I_{i}^{D}$ is the input of the method and $i$ represents the current frame. 

Specifically, each incoming depth frame is divided into a set of convex sub-instance-level segments using the geometry-based method described in \cite{furrer2018incremental}, based on the idea that real-world objects have overall convex surface geometries. For example, a chair instance belonging to the chair class undergoes further segmentation into components such as chair legs and chair back. At every depth image, surface normals are initially calculated, followed by a comparison of angles between adjacent normals to identify the edges of concave zones. This process leverages local pixel neighbourhoods to ascertain each pixel's local convexity. Additionally, the detection of significant depth discontinuities capitalizes on the large 3D distances between neighbouring depth map vertices. Ultimately, the 3D distance measure and surface convexity amalgamate to generate a set of convex sub-instance level segments $\left \{s_i^1,s_i^2,s_i^3,...,s_i^{j-1},s_i^j \right \}$ in the current frame $i$. 
\begin{equation}
 {\{s_i^1,s_i^2,s_i^3,...,s_i^{j-1},s_i^j\}} = \mathcal{F} (I_{i}^D)
\end{equation} where $j$ represent the $j$ th sub-instance level segments.
We denote the $p$ th semantic class and $q$ th instance in the $i$ th frame as: $\mathcal{O}_{i}^{p}$ and $\mathcal{N}_{i}^{q}$, obviously,  $ s_i^j \in \mathcal{N}_{i}^{q}$, $\mathcal{N}_{i}^{q}\in  \mathcal{O}_{i}^{p}$.

However, although the segments of all semantic classes are obtained, only relying on the geometric features of the segments themselves, it is not possible to complete class discovery through clustering. Next, we use these segments as carriers to query the corresponding semantic embedding $E_{i}^j$ and entropy $\varepsilon_i^j$ in the output of Embedding-NeRF. This process makes us not only obtained the segmentation fragments of each semantic class but also obtained the corresponding globally consistent features, which is of great benefit for novel class discovery.

\subsection{Feature Modulation}
After querying the corresponding features for sub-instance-level segments, we must incorporate these features into the segments to facilitate capturing more semantic information and class uncertainty information. The addition of these globally consistent features ameliorates issues of "over-segmentation" and "misclassification," ultimately yielding more accurate known and novel classes through clustering.

\noindent\textbf{Feature modulation}. Initially, we employ the $jth$ sub-instance-level segment in the $ith$ frame image to access the output of Embedding-NeRF, subsequently querying to obtain the semantic embedding and entropy attributable to this particular segment. Following this, in the feature modulation (in Algorithm 1), we concatenate the semantic embedding and entropy of the $jth$ segment, denoted as $E_{i}^j$ and $\varepsilon_i^j$, respectively, to generate the final comprehensive feature, $\mathcal{H}_i^j$, of the segment:
\begin{equation}
\mathcal{H}_i^j =  E_{i}^j \oplus \varepsilon_i^j
\end{equation} where $\oplus$ means concatenate operation, $\mathcal{H}_i^j$ means a high-dimensional vector, its dimension is $N \times H \times W \times (S+1) $.

\begin{algorithm}
    \caption{Feature Modulation}
    \label{alg:feature-aggregation}
    
    \begin{algorithmic}[1] 
        \State \textbf{Input:} a set of Convex Instance-level Segment $S$, Semantic Embedding $E$, Entropy $\varepsilon$.
        \State \textbf{Output:} Final segment feature $\mathcal{H}$
        
        \Function{Feature\_Modulation}{$S$, $E$, $\varepsilon$}
            \State $N \gets \text{length}(S)$
            \State $H, W \gets \text{shape}(S[0])$
            \State $S_{emb} \gets \text{shape}(E)[-1]$
            \State $\mathcal{H} \gets \text{zeros}((N, H, W, S_{emb}+1))$
            
            \For{$i = 0$ \textbf{to} $N-1$}
                \State $mask \gets S[i]$
                \State $embedding \gets E[mask]$
                \State $entropy \gets \varepsilon[mask]$
                
                \State $combined \gets \text{concat}((embedding, entropy), \text{axis}=-1)$
                
                \For{$h = 0$ \textbf{to} $H-1$}
                    \For{$w = 0$ \textbf{to} $W-1$}
                        \If{$mask[h, w] == 1$}
                            \State $\mathcal{H}[i, h, w, :S_{emb}+1] \gets combined[h, w]$
                        \EndIf
                    \EndFor
                \EndFor
            \EndFor
            
            \State $\mathcal{H} \gets \text{reshape}(\mathcal{H}, (N, -1))$
            \State \Return $\mathcal{H}$
        \EndFunction
    \end{algorithmic}
\end{algorithm}

\subsection{Markov clustering}
\noindent\textbf{Markov Clustering Based on Cosine Similarity}.we compute the cosine similarity \cite{tao2019multi} between sub-instance-level segments $s_ i^j$ base on it feature $\mathcal{H}_i^j$, the cosine similarity is a measure of similarity based on the cosine of the angle between two nonzero vectors of an inner product space. 
\begin{equation}
Similarity(s_i^m,s_i^n) = \frac{\mathcal{H}_i^m\mathcal{H}_i^n}{\left \| \mathcal{H}_i^m \right \| \left \| \mathcal{H}_i^n \right \| }   
\end{equation} where $m  \ne n$.

Through clustering, the process initially aggregates sub-instance level segments belonging to a single instance, effectively mitigating the "over-segmentation" issue resulting from geometric segmentation. Subsequently, instances corresponding to the same semantic class are clustered together, which not only resolves over-segmentation but also facilitates the discovery of novel classes. Specifically, we employ the Markov clustering algorithm(MCL) \cite{xu2005survey} because of the flexible number of clusters and computational cost. As we were unable to locate all clustering parameters in \cite{nakajima2019incremental}, we hand-tune these parameters until achieving optimal results for the kitchen\_0004 scene in the NYUv2 dataset, subsequently employing these settings (inflation = 12) across all scenes.

\begin{table*}
\centering
\caption{Assessment of Quantitative Outcomes for 9 Known and 4 Unknown Classes within the Replica Dataset: Semantic-NeRF and iLabel facilitate supervised segmentation of known classes via sparse label propagation, whereas NeurNCD not only segments known classes but also discovers novel classes.}
\label{tab:freq}
\begin{tabular}{ccccccc}
\toprule
Method & Label Propagation & known class mIoU & Novel class mIoU & Avg Acc & Total Acc \\
\midrule
 & Single Click & 50.1  & - & 84.7 & 80.5 \\
Semantic NeRF \cite{zhi2021place} & 1 \% & 68.2  & - & 82.7 & 84.5 \\
 & 5 \% & 76.5 & - & 86.3 & 87.1 \\
 & 10 \% & 80.9 & - & 88.3 & 89.1 \\
\addlinespace
 & 20 click & 48.0 &-  & - & - \\
iLabel \cite{zhi2022ilabel} & 40 click & 64.0 & - & - & - \\
 & 60 click & 72.0 &- & - & - \\
 & 80 click & 78.0 &- & - & -  \\
\midrule
Our & - & \textbf{81.3} & \textbf{50.6} & \textbf{89.1} &\textbf{89.7 } \\
\bottomrule
\end{tabular}
\end{table*}

\section{Experimental Evaluation}

\subsection{Datasets and Metrics}
\noindent\textbf{NYUv2}. We evaluate our proposed method on the NYUv2 \cite{silberman2012indoor} dataset. Following the official guide, we preprocess the entire dataset with MATLAB and use Open3D \cite{zhou2018open3d} to compute the camera poses. We train a separate Embedding-NeRF model for each scene, generating semantic embeddings and entropy. The official split of 654 images is used for testing. Images are resized to a resolution of 320 × 240 pixels in all experiments.

\noindent\textbf{Replica}. Replica \cite{straub2019replica} is a reconstruction-based 3D dataset containing 18 high-fidelity scenes with dense geometry, HDR textures, and semantic annotations. \citeauthor{zhi2021place} use the Habitat simulator \cite{savva2019habitat} to render RGB images, depth images, and semantic labels from randomly generated 6-DOF trajectories, mimicking hand-held camera motions. We evaluate our method's performance in discovering novel classes using their open-source simulated dataset. Images are resized to a resolution of 320 × 240 pixels in all experiments.

\noindent\textbf{Metric}. We use pixel classification accuracy (Acc.) and mean intersection over union (mIoU) as our metrics.

\subsection{Implementation Details}
\noindent\textbf{Pre-trained Model}. Utilizing ESANet\cite{seichter2021efficient} with a ResNet34 NBt1D backbone as our semantic segmentation network, we train it on the SUN RGB-D dataset\cite{song2015sun} for semantic embedding extraction, consisting of 5,285 RGB-D images. Specifically, by fine-tuning the dataset to include only 9 classes as known classes among the 13 classes defined in \cite{couprie2013indoor} and masking 4 classes as novel classes. 

For the NYUv2 Datasets, we use pre-trained ESANET to extract semantic embeddings from the RGB-D images. The selected classes and the entire classes are shown in Table 1. 

For the Replica dataset, we use the same pre-trained network to extract the semantic embedding. Since Semantic-NeRF is supervised and cannot discover novel classes. To ensure the fairness of the experiment, we compare NeurNCD with the sparse label propagation experiments in Semantic-NeRF. That is, in Semantic-NeRF, we apply single-click, 1\%, 5\% and 10\% pixel annotations to the 9 known semantic classes we defined previously, and generate semantic segmentation results through weak supervision. 

The above model is trained on a single 3090Ti GPU with 24GB memory. The batch size of rays is set to 1024 and the neural network using the Adam optimiser \cite{kingma2014adam} with a learning rate of 5e-4 for 200,000 iterations. The training time is approximately 8 hours, consistent with Semantic-NeRF training time but faster than the original NeRF (time > 24 hours).

\begin{figure}[!htb]
  \centering
     \includegraphics[width=0.95\linewidth ]{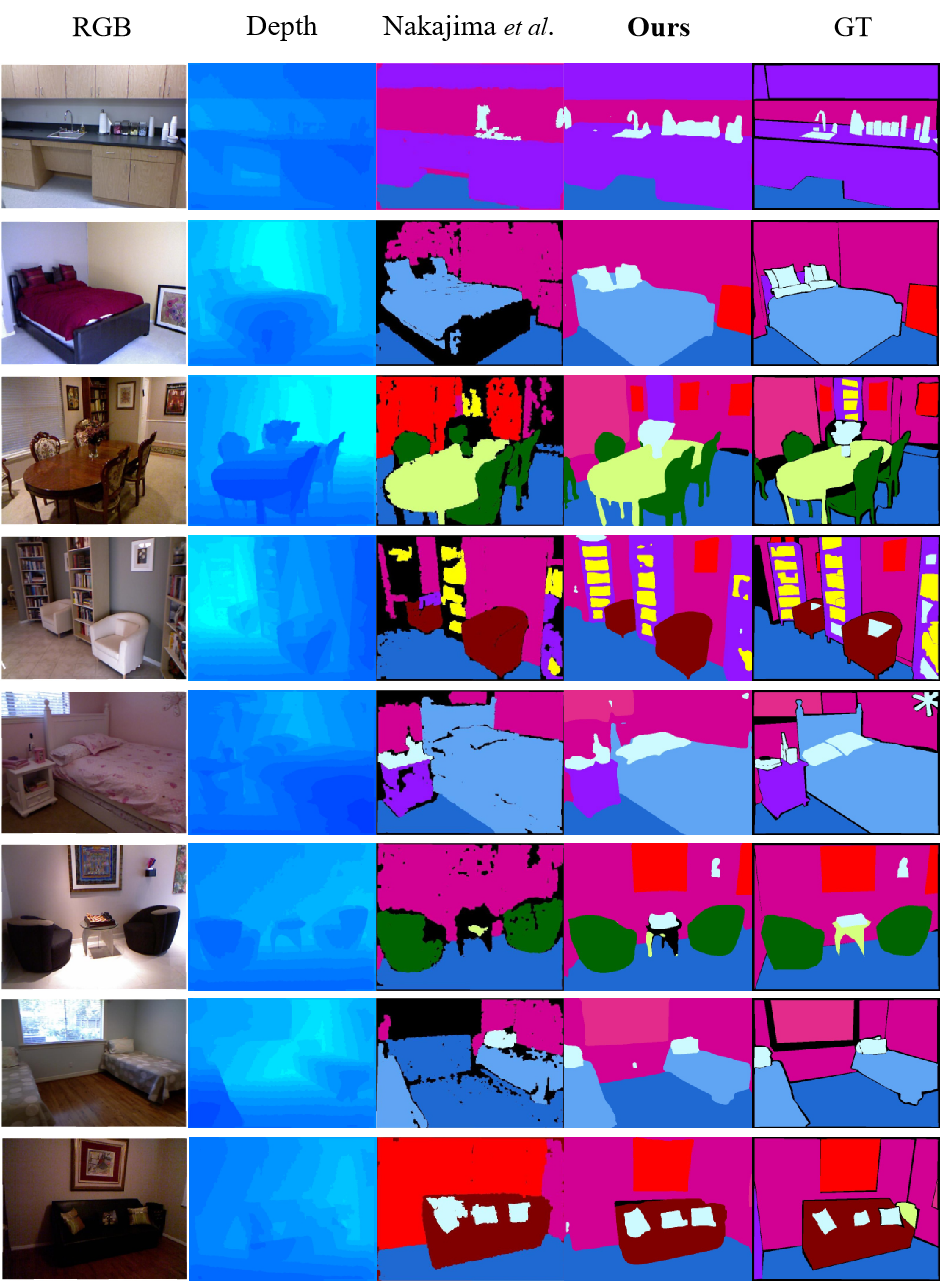}
   \caption{Results on the NYUv2 dataset. The third column is the result obtained by the method proposed by \citeauthor{nakajima2019incremental} and the fourth column is the result obtained by our method.}
   \label{fig:onecol}
\end{figure}
\subsection{Baselines}
As there are no previous works that use the neural radiation field to tackle the discovery of novel class problems, we compare our proposed method to the four most closely related approaches, i.e., two explicit representation methods\cite{nakajima2019incremental,tateno2015real} and two supervised implicit representation methods\cite{zhi2021place,zhi2022ilabel}.  

The explicit representation work was put forth by \citeauthor{nakajima2019incremental} we implement the method using the framework of \cite{blum2023scim}. Since we could not find all clustering parameters in \cite{nakajima2019incremental}, we use the parameter optimisation from \cite{blum2023scim} on the inflation and $\eta $ parameter of the MCL clustering for every scene.

Although no work has explored the important problem of discovering new classes in the open world with the help of neural radiation fields, in Semantic-NeRF\cite{zhi2021place} and iLabel\cite{zhi2022ilabel},  \citeauthor{zhi2021place} method for semantic segmentation via the sparse label propagation validated the potential of neural radiation fields to discover new classes, therefore, we also compared these two methods.

\subsection{Results}
Our experiments aim to demonstrate the effectiveness of the proposed method both statistically and subjectively. Firstly, we use the NYUv2 dataset for quantitative comparison, calculating the intersection over union (IoU) and presenting the results in Table 1.

Table 1 compares our method with two fully supervised methods and two unsupervised methods. Specifically, the fully supervised methods encompass a conventional semantic segmentation\cite{seichter2021efficient} and SLAM mapping for semantic segmentation\cite{nakajima2018fast}, while the unsupervised approaches include a state-of-the-art semantic mapping technique\cite{nakajima2019incremental} and a prior incremental 3D geometric segmentation method \cite{tateno2015real}, which served as inspiration for our work's geometric segmentation.

It becomes apparent that fully supervised methods are limited to predicting the nine classes in the training dataset and are incapable of uncovering novel classes. In stark contrast, our method significantly surpasses the other unsupervised techniques for both known and novel classes, achieving a mean IoU of 51.29. Although the unsupervised methods\cite{nakajima2019incremental,tateno2015real} can identify certain novel classes, their dependence on feature extraction and updating hampers their capacity to amalgamate multi-view visual features, thereby leading to incorrect segmentation and the generation of noise and outliers.

By capitalizing on Embedding-NeRF's feature fusion capabilities, our approach rectifies and supplements the "incomplete classes" and "outlier classes" arising from imprecise geometric segmentation. Quantitatively, our method enhances the mean IoU from 46.05 to 51.29 compared to the state-of-the-art technique. In the known class section, all classes in our results display significant improvement, whereas, in the unknown class section, three classes outperform state-of-the-art methods, albeit the picture class has no notable enhancement. This discrepancy stems from our geometric segmentation reliance on depth images alone, as opposed to the method in \cite{nakajima2019incremental}, which employs both depth and colour for segmentation, resulting in a marginally inferior performance for classes with suboptimal geometric features. 

In Table 2, we compare our approach with sparse label annotation propagation experiments from Semantic-NeRF \cite{zhi2021place} and iLabel \cite{zhi2022ilabel} as baselines to showcase the improvements in discovering novel classes in the Replica dataset. The pre-trained semantic segmentation network, as well as the known and new class settings, remain consistent with the NYUv2 dataset. Unlike Semantic-NeRF and iLabel's supervised semantic segmentation methods, our approach not only excels in known class segmentation but also discovers novel classes.

Semantic-NeRF achieves semantic segmentation using partial annotations consisting of Single Click or 1\%, 5\%, or 10\% of pixels per class within frames, while iLabel provides semantic segmentation results after 20, 40, 60, and 80 interactive clicks. Our method's segmentation of the nine known classes (mIoU=81.3) significantly outperforms the baseline (mIoU=80.9 and mIoU=78.0).

 \begin{table*}
 \large
\centering
 \caption{Ablation study for our method on NYUv2 dataset.}

\scalebox{0.80}{
\renewcommand{\arraystretch}{1.25}
\begin{tabular}{ccccc|ccccccccc|cccc|c} 
\hline
\multicolumn{5}{c|}{Components}     & \multicolumn{9}{c|}{classes in training dataset}                                                                            & \multicolumn{4}{c|}{novel classes}       & mIoU       \\ \hline
GS           & PSSN        & EP    & SE    & MCL         & bed          & book           & chair          & floor                   & furn.       & obj.     & sofa      & table       & wall         & ceil.          & pict.          & TV                      & wind.          & ~             \\ \hline

 $\checkmark$ &   &     &     &                  & ~48.77         & 20.76          & ~35.23         & 49.89          & ~45.25         & ~20.86         & ~40.18         & ~38.22         & 51.80          & 27.76          & 15.93          & ~45.88         & ~39.23         & ~36.91~           \\
 & $\checkmark$ &                            &               &                 & ~51.67         & ~26.19         & 41.55          & ~50.79         & ~54.24         & ~22.18         & ~~43.75        & ~44.86         & ~57.88         & ~-             & -~             & -~             & -~             & ~-~           \\
$\checkmark$                     & $\checkmark$     & $\checkmark$         &  & $\checkmark$ & 53.55          & 27.10          & 44.79          & 55.93          & 59.07          & 25.12          & 45.98          & 45.19          & 59.25          & 31.89          & 19.64          & 49.24          & 42.82          & ~43.04~           \\
$\checkmark$                    & $\checkmark$                   & $\checkmark$                            & $\checkmark$ & $\checkmark$ & \textbf{69.23} & \textbf{29.82} & \textbf{58.63} & \textbf{69.67} & \textbf{60.11} & \textbf{32.18} & \textbf{58.86} & \textbf{48.25} & \textbf{69.28} & \textbf{31.92} & \textbf{25.59} & \textbf{59.38} & \textbf{53.95} & \textbf{~51.29~}  \\ 
\hline

\end{tabular}}

\end{table*}

\begin{figure}[htb]
  \centering
   \includegraphics[width=\linewidth]{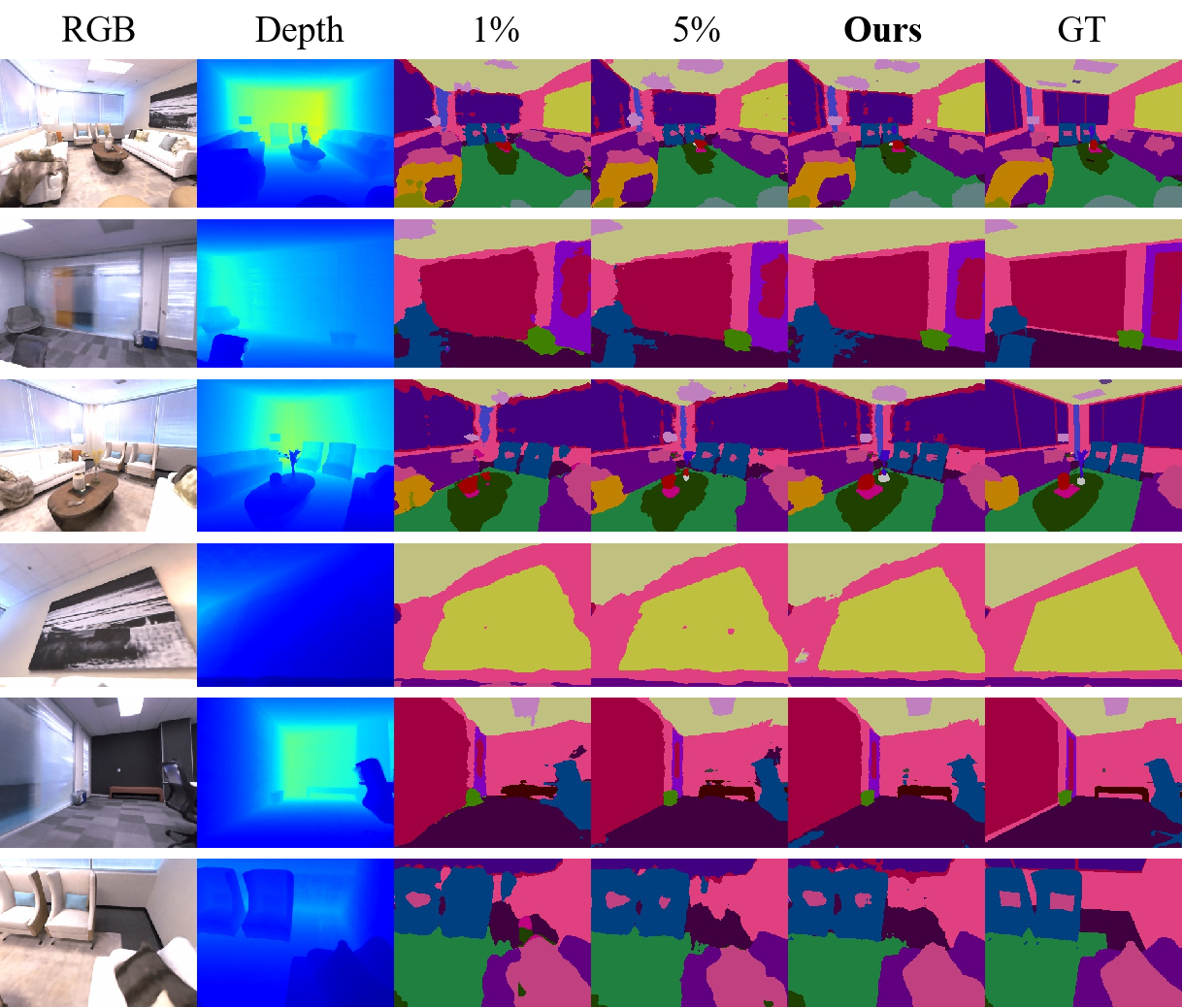}
   \caption{Results on the Replica dataset. The third and fourth columns are the results of sparse annotation 1\%, 5\%, column 5 is the result of our method, and column 6 is the semantic ground truth.}
   \label{fig:onecol}
\end{figure}

\subsection{Ablation Studies}
In this section, we assess the influence of various components, such as the geometric segmentation (GS), pre-trained semantic segmentation network(PSSN), semantic embedding (SE), entropy (EP), and Markov clustering (MCL), with ablation study results presented in Table 3.

Comparing GS and PSSN, we observe that the pre-trained model's segmentation performance on known categories significantly surpasses that of pure geometric segmentation. However, the pre-trained model fails to identify unknown categories. While geometric segmentation can detect unknown categories, it assumes that real-world objects exhibit overall convex surface geometry, leading to suboptimal segmentation results for objects with poor convexity and causing over-segmentation. This issue necessitates a reliance on clustering algorithms to associate segments of the same category.

Examining the contributions of EP and SE in feature modulation, we conduct ablation comparison experiments for two cases: EP and EP +SE. Results reveal that incorporating only entropy features into segments yields inferior outcomes while adding semantic embedding improves performance. This is because Entropy, being discrete and lacking spatial consistency, can only assign characteristics to known classes, offering minimal aid in discovering unknown classes. Conversely, the semantic embedding derived from Embedding-NeRF demonstrates spatial consistency and continuity.

Moreover, the feature modulation module facilitates the assignment of entropy and embeddings to both known and unknown classes. Consequently, following clustering, sub-instance-level segments with matching features and embeddings can effectively accomplish the segmentation of known classes and the discovery of novel classes.

\section{Conclusion}
We investigate discovering novel classes in open-world settings and propose a powerful solution named NeurNCD. We develop exquisite designs like Embedding-NeRF with KL divergence, feature query and modulation, and unsupervised clustering. Whether compared with traditional explicit representation methods or with supervised implicit representation methods, Our method shows superior quantitative and qualitative results in both known class segmentation and novel class discovery.


\newpage
\bibliographystyle{ACM-Reference-Format}
\balance
\bibliography{sample-sigconf}

\appendix

\end{document}